\documentclass[letterpaper, 10 pt, conference]{ieeeconf}  %

\IEEEoverridecommandlockouts    %
\usepackage[utf8]{inputenc}
\DeclareUnicodeCharacter{2212}{-}
\usepackage{amsmath,amsfonts}
\usepackage{algorithmic}
\usepackage{algorithm}
\usepackage{array}
\usepackage[caption=false,font=normalsize,labelfont=sf,textfont=sf]{subfig}
\usepackage{textcomp}
\usepackage{stfloats}
\usepackage{url}
\usepackage{verbatim}
\usepackage{graphicx}
\usepackage{multirow}
\usepackage{glossaries}
\makeglossaries
\usepackage{pgfplots}
\usepgfplotslibrary{groupplots,dateplot}
\usetikzlibrary{patterns, shapes.arrows}
\pgfplotsset{compat=newest}
\newcommand{\edited}[2]{{}{#2}}

\usepackage{siunitx}
\usepackage{pgf}
\usepackage{siunitx}
\usepackage[nocompress]{cite}
\usepackage[colorlinks=true]{hyperref}

\newacronym{cot}{CoT}{Cost of Transportation}
\newacronym{rl}{RL}{reinforcement learning}
\newacronym{RoM}{RoM}{Range of Motion}

\title{\LARGE \bf LEVA: A high-mobility logistic vehicle with legged suspension}

\author{
    Marco Arnold${}^{*}{}^{1}$, %
    Lukas Hildebrandt${}^{*}{}^{1}$, %
    Kaspar Janssen${}^{*}{}^{1}$, %
    Efe Ongan${}^{*}{}^{1}$, %
    Pascal Bürge${}^{2}$, \\%
    Ádám Gyula Gábriel${}^{1}$, %
    James Kennedy${}^{1}$, %
    Rishi Lolla${}^{1}$, %
    Quanisha Oppliger${}^{2}$, %
    Micha Schaaf${}^{2}$, \\%
    Joseph Church${}^{1}$, %
    Michael Fritsche${}^{1}$, %
    Victor Klemm${}^{1}$,  %
    Turcan Tuna${}^{1}$, %
    Giorgio Valsecchi${}^{1}$, %
    Cedric Weibel${}^{1}$,\\ %
    Michael Wüthrich${}^{2}$, %
    Marco Hutter${}^{1}$ %
    \thanks{This project has received funding from the European Union’s Horizon Europe Framework Programme under grant agreement No 101070596 (euROBIN)}
    \thanks{${}^{1}$Robotic Systems Lab, ETH Z\"urich, 8092 Z\"urich, Switzerland, 
    ${}^{2}$Zurich University of Applied Sciences,
    ${}^{*}$These authors contributed equally.}%
    \thanks{This paper has been accepted for publication at the 2025 IEEE International Conference on Robotics and Automation (ICRA).}
}

\begin{document}

\maketitle
\thispagestyle{empty}
\pagestyle{empty}

\begin{abstract}
The autonomous transportation of materials over challenging terrain is a challenge with major economic implications and remains unsolved. This paper introduces \textit{LEVA}, a high-payload, high-mobility robot designed for autonomous logistics across varied terrains, including those typical in agriculture, construction, and search and rescue operations. \textit{LEVA} uniquely integrates an advanced legged suspension system using parallel kinematics. It is capable of traversing stairs using a \gls{rl} controller, has steerable wheels, and includes a specialized box pickup mechanism that enables autonomous payload loading as well as precise and reliable cargo transportation of up to \textbf{85 kg} across uneven surfaces, steps and inclines while maintaining a \gls{cot} of as low as 0.15. Through extensive experimental validation, \textit{LEVA} demonstrates its off-road capabilities and reliability regarding payload loading and transport. 
\end{abstract}

\section{Introduction}
\label{introduction}
The logistics industry has experienced significant advancements in automation in the past decades, particularly with the deployment of wheeled robots optimized for the efficient transport of heavy payloads within structured environments like warehouses. These robots excel on flat surfaces, where their efficiency and payload handling are unmatched. However, their application is limited in less structured environments featuring stairs, steps, and uneven terrains, where their locomotion capabilities are diminished, or deployment becomes infeasible.

Legged robots, characterized by their superior agility and ability to adapt to challenging terrains, offer potential solutions for these less accessible areas. Despite this, they generally lag behind wheeled robots in both load-carrying capacity and operational efficiency. Although recent improvements have enhanced their agility \cite{doi:10.1126/scirobotics.adi7566}, energy efficiency \cite{10246325}, and speed by adding wheels \cite{8642912}, these advancements have not yet provided a scalable solution for autonomous logistics across rougher terrains. 

For these use cases, a wheeled-legged robot with the ability to autonomously handle heavy payloads, including loading and unloading them on its own, could substantially broaden the operational scope of these systems and set a robot apart from existing solutions.

Such an integrated approach would be particularly beneficial in scalable logistics applications crucial for sectors like agriculture, last-mile delivery, and intralogistics. These sectors not only demand high autonomy and payload capacity but also require robust terrain adaptability. 
\begin{figure}
    \centering
    \includegraphics[width=1\linewidth]{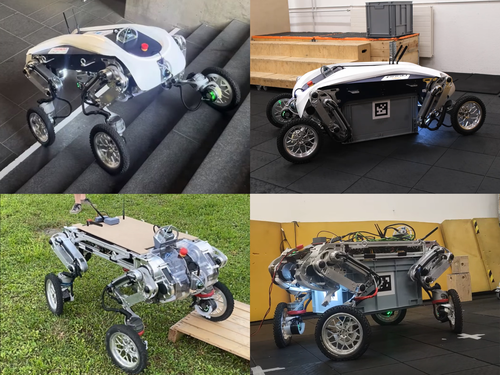}
    \caption{The high payload robot \textit{LEVA}, shown in multiple environments}
    \label{fig:SystemOverview}
\end{figure}

\section{Related work}
Heavy-payload, off-road logistics demand a system capable of overcoming various obstacles, typically addressed with three key designs: wheels, tracks, and legs.

Indoor environments see purely wheeled systems, like the C-matic 06 \cite{LindeCMATIC}, achieving very low \gls{cot}, high speeds, and extremely heavy payloads, making them nearly unbeatable in such settings. However, they are restricted to completely flat surfaces. Wheeled systems like the GPR-4400 from Ambot \cite{AmbotGRP4400}, while capable of navigating small steps up to the wheel radius, still suffer from significant tilting and shaking of the body and payload, limiting their overall mobility.

Tracked vehicles, such as the KOMODO02 from jcrobot~\cite{JCSKomodo}, share many advantages and drawbacks with wheeled robots but often perform better off-road due to their larger ground contact area. Nevertheless, tracks employ skid-steering and, therefore, operate with reduced efficiency when needing to turn often.

Legged systems focusing on payload, still in the research phase, exemplified by Barry~\cite{10246325} and BigDog~\cite{Raibert2008BigDogTR}, offer the greatest mobility, navigating most urban and off-road terrains effectively. Barry is tailored for efficient and heavy payload handling, while BigDog supports substantial loads with its hydraulically powered legs.

In logistics applications, maximum payload capacity, velocity, cost of transport and operational times are the most relevant metrics for comparing solutions. Table~\ref{table:CoTleg} lists these values for the robots introduced above. Despite these capabilities, none of the mentioned systems —wheeled, tracked, or legged— include dedicated features for fully autonomous payload management, relying instead on human operators or auxiliary systems for loading. Robots that do offer such autonomy, like Handle from Boston Dynamics~\cite{RobotsGuideHandle}, Digit from Agility Robotics~\cite{Digit} or the robot Alphred~\cite{Alphred}, typically sacrifice size, efficiency, or speed compared to the aforementioned examples.

\vspace{5mm}
 \begin{table*}[ht]
 \centering
 \scriptsize
 \caption{Overview of key metrics for different payload handling robots}
 \resizebox{\textwidth}{!}{\begin{tabular}{lllccccccccccc}
 \hline
 \textbf{Name} & \textbf{Type} & \textbf{Year}  &  \multicolumn{2}{c}{\textbf{Payload [Kg]}} & \textbf{Mass} & \textbf{Payload to} & \multicolumn{2}{c}{\textbf{Velocity [m/s]}} & \textbf{Power} & \textbf{Autonomy} & \textbf{CoT}
 \\
  &  &  & \textbf{Ref.} & \textbf{Max} & \textbf{[kg]} &  \textbf{Mass Ratio}\(^a\) & \textbf{Ref.} & \textbf{Max.}&  \textbf{[W]} & \textbf{[h]}  &  
                              \\ \hline
  C-Matic 06 \cite{LindeCMATIC}& wheeled logistics& 2022& 70 & 600 & 145& 4.13& 1.5& 1.5  &-& 23.44 & 0.08 \(^a\) (0.02 with max. payload)\\
  GRP 4400 \cite{AmbotGRP4400}& wheeled logistics robot& 2015& - & 250 & 250& 1.0 & -& 5.4  & - & - & - \(^a\)\\
   KOMODO 2 \cite{JCSKomodo}& tracked logistic robots& 2022& 100 & 100 & 110& 0.9 & 1.6& 1.6  &1000& - & 0.52 \(^a\)\\
 BigDog  \cite{Raibert2008BigDogTR} \(^e\)           & quadruped (hydr.)         & 2008                       & 50 & 154          & 109                & 0.46-1.4                              & 1.1           & 3.1         & 11000\(^b\)        & 2.5          & 15\\
 Barry \cite{10246325}       & quadruped         & 2022                 & 50                    & 90            & 48                 & 1.1-2                               & 1.4           &2.0         & 370     & 2.5\(^a\)      & 0.7       \\ 
 
 LEVA (this paper)     & wheeled legged         & 2024                          & 70                    & 100  \(^c\)           & 85            & 1.0-1.2                               & 1.3&2.0         & 472&  1.5& 0.23 (on legs), 0.15 (on bump-stops)\\
  \hline
 \end{tabular}}
  \text{\(^a\)Estimated from other parameters in this table or from work/data sheets cited \(^b\)Maximum Value, not used in the estimation of other parameters }\\
 \text{
 \(^c\)Theoretical, from design specifications or max. joint torques but not demonstrated in joint torques. Not used in the estimation of other parameters}
 \text{
 \(^d\)Not including power consumption due to computation}
\text{\(^e\)Data taken from paper featuring Barry \cite{10246325}}
\label{table:CoTleg}
\end{table*}
In this paper, we present \textit{LEVA}, shown in Fig.~\ref{fig:SystemOverview}, a novel robotic platform designed to enable off-road logistics. Our main contributions can be summarized as follows: 
\begin{itemize}
    \item An active suspension based on parallel kinematics and actuated steerable wheels allowing for driving and stepping. 
    \item A dedicated mechanism and algorithms for fully autonomous payload pickup and placement.
    \item The successful deployment of an \gls{rl}-based controller for a parallelogram leg with a steerable wheel to traverse steps, stairs, and natural grounds.
\end{itemize}

\begin{figure*}[t]
      \centering
      \begin{minipage}[b]{0.275\textwidth}
        \includegraphics[width=\textwidth]{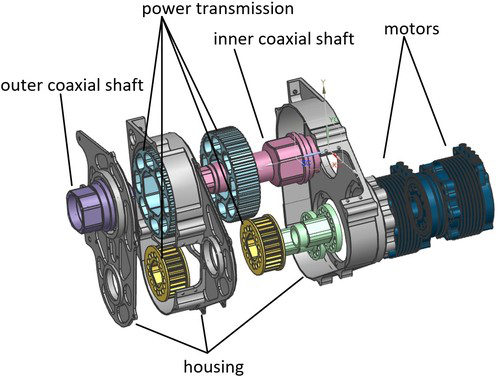}
        \caption{Explosion of shoulder gearbox, belts not included}
        \label{fig:shouldergearbox}
      \end{minipage}
      \hfill
      \begin{minipage}[b]{0.45\textwidth}
        \includegraphics[width=\textwidth]{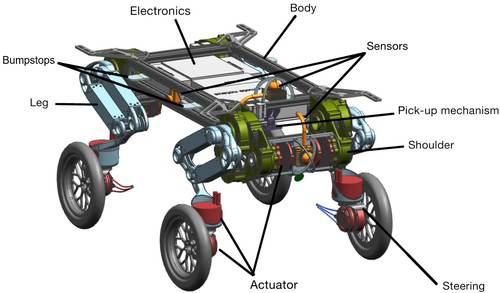}
        \caption{Mechanical system overview, shell not included. \edited{}{Sensors are further explained in Fig.~\ref{fig:Sensors}}.}
        \label{fig:CADOverview}
      \end{minipage}
      \hfill
      \begin{minipage}[b]{0.26\textwidth}
        \includegraphics[width=\textwidth]{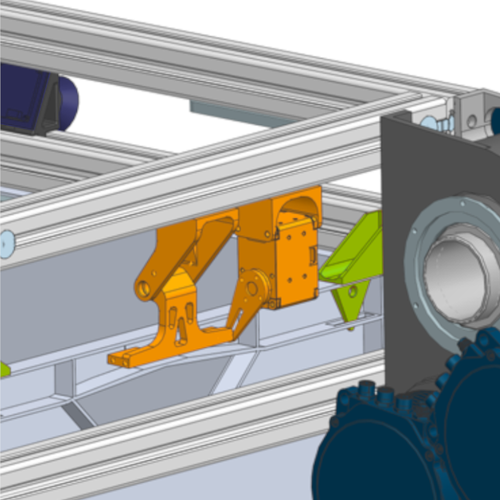}
        \caption{Pickup mechanism; hook system (orange) and alignment pins (green) are shown.}
        \label{fig:pickup}
      \end{minipage}
    \end{figure*}
\section{System Design}
\label{SysDesign}
\subsection{System Overview}
\textit{LEVA} is \SI{1.2}{\meter} long and \SI{0.75}{\meter} wide with an adjustable rolling height of \SI{0.6}{\meter} to \SI{0.9}{\meter}. These dimensions were determined to be able to navigate through most indoor spaces, including doors and corridors that conform to the DIN 18101 and 18040 standards. They were a critical design requirement for the leg and shoulder structures. The robot, weighing \SI{85}{\kilogram}, has a verified payload-to-weight ratio of one.

To address the challenge of autonomous payload loading and unloading, \textit{LEVA} employs a strategy that allows for the secure attachment of cargo beneath the robot through specialized gripping and aligning interfaces. This design enables \textit{LEVA} to autonomously lower itself over the cargo, utilizing its legs to facilitate loading and unloading without external assistance. Currently, this system is optimized for handling standardized EuroBoxes of the size \SI{0.6}{\meter} $\times$ \SI{0.4}{\meter} and varying height, which can be found in various applications due to their widespread use and versatility.

\textit{LEVAs} four legs have four degrees of freedom each, allowing for planar leg movement, and include an additional wheel with a steering axis. This design enables a practical balance between the agility of legged movement and the smooth omnidirectional rolling typical of wheeled transport, while remaining slim. The legs are arranged in an "X" configuration close to the robot’s body to conform with \textit{LEVAs} indoor usability.
\subsection{Leg Design}
\label{legdesign}
A high payload-to-torque ratio is fundamental to \textit{LEVAs} leg design and is essential for its lifting and climbing capabilities with heavy loads. The robot's legs are each comprised of a four-bar linkage system that is driven coaxially from the shoulder, enabling its two actuators to drive the linkage coaxially and thus work in tandem. We chose this topology because its mechanical advantage requires relatively small torques for producing high vertical forces in specific parts of the range of motion, particularly in zones crucial for overcoming obstacles, where stepping intensifies the load on individual legs.

\begin{figure} [H]
    \centering
    \includegraphics[width=1.\linewidth]{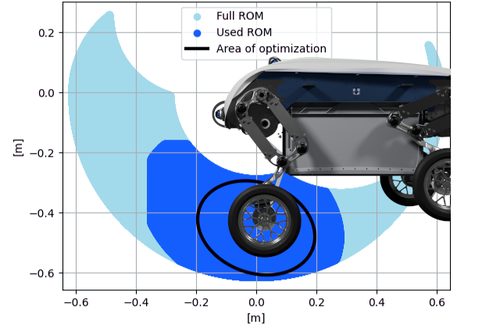}
    \caption{\textit{1.} Maximal possible \gls{RoM} in light blue, \textit{2.} \gls{RoM} used \edited{in operation}{during box pickup and locomotion} in dark blue, \textit{3.} \edited{Area optimized for low motor torques is shown.}{Area optimized for low motor torques during stair climbing is shown inside the black ellipse.} 
    }
    \label{fig:rom_leg}
\end{figure}

The different lengths of the leg were estimated by using static equilibrium calculations and a grid search over geometric leg parameters to minimize the sum of squared torques resulting from a vertical force applied to every foot position in a \gls{RoM} (Fig.~\ref{fig:rom_leg}) sufficient for overcoming the obstacles mentioned in Section~\ref{introduction}. In addition to the limb lengths, the optimization also included the positions of the rotational axes of the actuated limbs. Aligning these axes improved the range of motion and eliminated singularities in critical leg positions.

Parallelogram legs are known to be inefficient during walking due to antagonistic power consumption \cite{Antagonism}. This principle also applies to \textit{LEVAs} wheeled locomotion, with static legs, resulting in power consumption mainly through motor losses. To increase the efficiency, \textit{LEVA} features a set of bump stops (see Fig.~\ref{fig:CADOverview}) that introduce a resting position where the shoulder actuators do not draw power.
\subsection{Shoulder Design}
The upper limbs of \textit{LEVAs} legs are actuated coaxially from the shoulder. To achieve this in a way that does not protrude outside of the predetermined dimensions, \textit{LEVA} features a dedicated shoulder gearbox that not only realizes the coaxial drive system but also amplifies the motor torques via two belt drives (see Fig.~\ref{fig:shouldergearbox}). This design offers several advantages: the motors are not mounted directly on the leg, preventing axial shocks from being transmitted to the actuators; allowing for torque amplification and looser tolerances compared to gear transmissions. We chose state-of-the-art Maxon HEJ 90-48-140 actuators as the shoulder drives due to the following reasons: The drive makes use of a planetary gearbox and benefits therefore of a high torque transparency coupled with a high torque-density ratio. 

\subsection{Steering Design}
\label{steering}
To navigate flat environments efficiently and swiftly, \textit{LEVA} is equipped with four wheels, each with a radius of \SI{14}{\cm}, and an omnidirectional drive system that enables both efficient travel and precise maneuvering. Those objectives are achieved through dedicated steering actuators composed of U-TZ90-1A from Unity Drive Systems at each wheel. 

To keep the design compact, the steering axis is offset from the wheel's contact point with the ground (see Fig.~\ref{fig:CADOverview}). To minimize static holding torques for the motor which would reduce efficiency, the steering axis is angled perpendicular to the ground in the robot's typical driving and stationary positions.
\subsection{Body Design} 
\label{Body}
The body of \textit{LEVA} is closely designed around the EuroBox to minimize space usage and ensure a slim profile. The body structure is as wide as the payload container, enclosing it only on the short sides. Additionally, the shoulders are strategically mounted at the front and back, with only the legs extending beyond the width of the EuroBox.
\subsection{\edited{}{Box Pickup Design}} 
The box pickup mechanism is tightly integrated into the body's design, as seen in Fig.~\ref{fig:pickup}. It consists of a hook actuated by a servo between the front and rear shoulder pairs. The hooks are hinged in a way that close to no loads are projected on the servo motors when a box is connected. Additionally, the body design features multiple alignment features (slanted surfaces), allowing for passive mechanical alignment of the payload relative to the robot. They can compensate for up to $\pm$ \SI{3}{\cm} of lengthwise and $\pm$ \SI{1}{\cm} of crosswise alignment error \edited{}{mainly due to perception and control inaccuracies}.

\section{Control System Overview}
\label{ContOverview}
For precise and efficient transport on quasi-flat terrain, an inverse kinematics-based controller is utilized, shown in Section~\ref{RollingCont}. In addition, to overcome challenging obstacles, an \gls{rl}-based perceptive locomotion controller is trained \ref{RLCont}; together, these controllers allow for efficient mobility.

For navigation and perception, a LiDAR-inertial odometry system is used in conjunction with cameras. A high-level planner using the perception and locomotion systems enables reliable box pickup \ref{BoxPick}.
\subsection{Sensors and State Estimation}
\subsubsection{Odometry Estimation}
\textit{LEVA} has two forward-facing Livox Mid-360 LiDARs (see Fig.~\ref{fig:Sensors}). The point cloud generated by the two LiDARs is used in conjunction with FAST-LIO2~\cite{xu2021fastlio2fastdirectlidarinertial} \edited{in order}{in order to} get LiDAR-inertial-based pose estimation. The obtained odometry estimates and the fused, deskewed point cloud are then used to generate a 2.5D elevation mapping~\cite{miki2022elevation} of the surroundings. 
\subsubsection{EuroBox Detection}
\label{Eurobox_detection}
\begin{figure}
    \centering
    \includegraphics[width=1.0\linewidth]{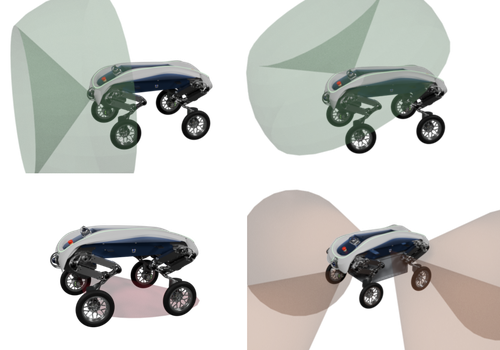}
    \caption{Overview of the fields of view of \textit{LEVA}'s sensors. \textit{Top left:} front facing LiDAR. \textit{Top right:} upwards facing LiDAR. \textit{Bottom left:} downwards facing camera for box perception. \textit{Bottom right:} outwards facing cameras on each of the four sides of the robot.}
    \label{fig:Sensors}
\end{figure}

For detecting the EuroBox, \textit{LEVA} includes 5 monocular cameras (Arducam B0385, as shown in Fig.~\ref{fig:Sensors}). Images obtained from the monocular cameras are fed into a parallelized apriltag-detection~\cite{apriltag} pipeline. The detection pipeline uses a low-pass filter in world coordinates to smooth the data and reduce the effects of outliers. After low-pass-filtered,  the detection is stored in a buffer.

 This buffering scheme enables \textit{LEVA} to be robust against perception failures. In the event of an apriltag-detection failure, the last detected box coordinate in the world frame, in conjunction with the LiDAR-inertial odometry module, is used to continue operation. This approach allows \textit{LEVA} to reliably track the EuroBox and execute pickup manoeuvres. 

 \begin{figure*}[t]
     \centering
     \includegraphics[width=\linewidth]{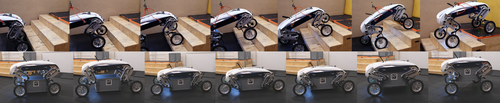}
     \caption{Stair climbing procedure (top, total \SI{6}{\second}) and Box pickup procedure (bottom, total \SI{25}{\second})}
     \label{fig:BoxAndStair}
\end{figure*}
\subsection{Rolling Controller}
\label{RollingCont}
The controller uses inverse kinematics to set the shoulder actuator positions from the desired robot height and stance. The steering angles and wheel speeds are calculated from normal and tangential velocities from linear/angular velocity commands, including wheel speed compensation to avoid slippage when steering (discussed in Sub-section~\ref{steering}). Since cables for the wheel motor result in a limited steering range, certain steering angles further away from the limits allow for higher control authority. In the case of multiple steering angle solutions, the shortest path with a bias towards a high control authority angle from the current angle is chosen. This enables continuous driving along smooth trajectories while avoiding running into steering limitations during significant direction changes.

\subsection{RL Controller}
\label{RLCont}
\subsubsection{Simulation}
The simulator utilized in this work, ORBIT~\cite{mittal2023orbit}, uses the PhysX 5.3 engine. This physics engine has no support for joints in closed kinematic chains with reduced coordinate representations. Thus, the loop is broken by simulating the final joint as a force joint.
This introduced simulation unreliability and created concerns about simulation accuracy. \textit{LEVA} requires an accurate friction model during training, but the underlying physics engine cannot create realistic tire slip friction forces, thus causing unrealistic steering behavior. To mitigate these problems, three main methods were developed:
\begin{itemize}
    \item To ensure the system dynamics were captured properly, a controller deployment with the rolling controller~\ref{RollingCont} was performed, during which actuator commands and measurements of joint positions and joint velocities were collected. Then, the collected actuator commands were replayed in the simulation environment, and the same measurements were collected.
    To ensure the reliability of the simulation, the data was compared to the real-world experiment and used for system-identification.
    \item A penalty was introduced to penalize behavior that resulted in parallelogram angles approaching the leg limits. This was done primarily to reduce the probability of system damage and secondarily to reduce simulation instability that is caused by the force joint requirement.
    \item To reduce unrealistic tire slippage, wheel velocities in the axial direction were penalized.
\end{itemize}

\subsubsection{Training}
For \gls{rl} training, the training pipeline used by \cite{pmlr-v164-rudin22a} was modified to fit the topology of the system, with RSL-RL~\cite{rsl_rl} chosen for algorithms. The teacher-student approach from \cite{Lee_2020} with an MLP teacher network and a GRU student network was selected for the architecture. For the training curriculum, the environments were separated into two classes: a first class with stair terrains of varying dimensions and a second class with quasi-flat terrains. The commands were sampled with system limitations in mind, i.e. \textit{LEVA} cannot achieve omnidirectional movement on stairs but is capable doing so on quasi-flat terrains. To use the binary nature of this separation, the terrain-boolean-based strategy from \cite{chamorro2024reinforcementlearningblindstair} was adopted.
\paragraph*{Noiseless Mixing}
A noiseless mixing-based technique was used to change the action distribution with more bias on the mean of the outputs compared to a Gaussian Distribution as action noise. During training, 10\% of the population was randomly selected and their respective output noise standard deviation was set to zero, i.e. made deterministic. This heuristic allowed for a better exploitation of the swerve steering mechanism. \edited{}{ Proving the effectiveness of the noiseless mixing requires further testing and currently presented as an heuristic.}

With these system-specific modifications to the training process, \textit{LEVA} is capable of achieving locomotion in diverse terrains and omnidirectional motion in quasi-flat terrains as illustrated in Fig.~\ref{fig:BoxAndStair}. The RL controller's precision is inferior to that of the rolling controller described in Section~\ref{RollingCont} when it comes to changing the height and the leg width, increasing the precision of the RL controller is to be investigated in the future.

\subsection{Box Pickup}
\label{BoxPick}
\begin{figure}
    \centering
    \includegraphics[width=1.\linewidth]{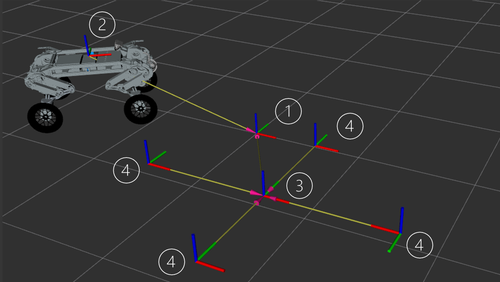}
    \caption{Box pickup frames and points: \textit{1.} World frame, \textit{2.} Robot frame, \textit{3.} Box frame, \textit{4.} Approach points}
    \label{fig:BoxAndStair_cos}
\end{figure}

The box pickup is coordinated by a high-level state machine which can be initialized when a EuroBox is detected by the perception subsystem mentioned in Section~\ref{Eurobox_detection}. The maneuver itself consists of three phases. In \textbf{phase I}, the closest of the four box approach points (Fig.~\ref{fig:BoxAndStair_cos}) is approached in a straight line while correcting the rotation of the robot relative to the box. Then, \textit{LEVA} moves over the box in \textbf{phase II}. When location and orientation are sufficiently precise, \textbf{phase III} starts (Fig. \ref{fig:BoxAndStair}), consisting of lowering the body while continuously correcting for displacement in the lengthwise direction, gripping the box, and standing up again.
\section{Experimental Verification}
\label{Experiment}
In this section, we describe the methods and results from evaluating \textit{LEVAs} payload transport, rough and uneven terrain traversal, autonomous payload handling, and efficient transportation performance.

\subsection{Terrain}
To validate \textit{LEVAs} terrain adaptability, we tested the system in different environments, some of which can be seen in Fig.~\ref{fig:SystemOverview}. These tests mainly included ramps with inclines of up to \SI{30}{\deg}, different surfaces like grass, gravel, and asphalt, and the traversal of small obstacles like curbstones or palettes up to a height of \SI{15}{\cm}. For all of these terrains, the rolling controller introduced in Section~\ref{RollingCont} was sufficient. During deployment, an operator can actively change between different stiffness modes to increase the compliance of the leg actuators and, thus, the terrain-adaptability in rougher environments.

The \gls{rl} controller has proven to be capable of traversing different stairs, shown in Fig.~\ref{fig:BoxAndStair}, indoor and outdoors.
\subsection{Box pickup}
The box pickup system, as outlined in Section~\ref{BoxPick}, was rigorously tested through 50 consecutive pickup and drop-off maneuvers with payloads ranging from 0 to \SI{70}{\kilogram}. The robot's starting positions varied, allowing it to approach and pick up the boxes both lengthwise and from the side, fully leveraging its omnidirectional capabilities. 

No significant correlation was observed between failures and the weight of the payloads. Failures during the maneuvers were primarily due to systemic issues rather than the pickup system. Specifically, actuator connection issues triggered a failure on three occasions, and camera connection issues caused another three failures. There were also two instances of slight misalignment during pickups; however, these did not prevent the successful completion of the maneuvers, validating the effectiveness of the alignment features introduced in Section~\ref{Body}. Despite these issues, the tests achieved an overall success rate of 86\%. When excluding failures not directly related to the maneuver itself, the success rate improves to 97.7\%. 
\subsection{CoT experiment}
\begin{figure}
    \centering
    \resizebox{1.\linewidth}{!}{\input{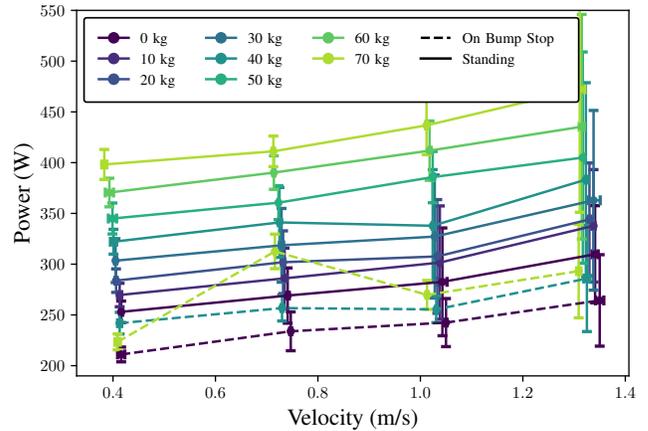}}  
    \caption{Plot showing \textit{LEVAs} power consumption for different payload masses and comparing standing and sitting on bump stops.}
    \label{Powertomass}
 \end{figure}

\gls{cot} is critical for assessing the efficiency of legged robots, especially for \textit{LEVA}, whose primary function is cargo transport. To analyze \gls{cot}, we used the established formula \ref{cot_eq}  

\begin{equation}
COT = \frac{P_{\text{Est}}}{g \cdot m_{\text{Tot}} \cdot v}, \label{cot_eq}
\end{equation}

where $v$ represents the base velocity, $g$ is the acceleration due to gravity, \edited{}{ $ m_{\text{Tot}}$ is the total mass of the robot including the payload} and $P_{\text{Est}}$ is the measured total power consumption.
To measure the total power consumption of \textit{LEVA}, we used an external calibrated bidirectional power supply, EPS 9200-1403 U, rated for \SI{10}{\kW}. The tests where conducted with the battery still installed but electrically disconnected and included the consumption of all electrical components, including actuators and two PC's. The total power consumption during the experiments is shown in Fig.~ \ref{Powertomass}. The \SI{200}{\W} idle power consumption is mainly due to the high amount of compute power available on demand.

During locomotion on flat ground, \textit{LEVA} does not need to actively use its legs for movement and, therefore, does not change its wheelbase. Consequently, the base velocity was determined as an average of the individual wheel velocities. These were gathered by combining wheel actuator rotational speeds from their encoders with a tire radius of \SI{14}{\cm}. 

To replicate real-world conditions, the experiments were conducted with \textit{LEVA} positioned at a predetermined height and stance, ensuring that it does not rest on its bump-stops. This configuration optimizes the leg actuator torques while allowing sufficient leg movement for damping in uneven terrains and accommodating high accelerations without the risk of tipping over. The primary aim of these experiments was to provide a fair comparison of \textit{LEVAs}’s Cost of Transport with quadrupeds.

The total weight in the subsequent experiments always included the total weight of the robot of \SI{85}{\kilogram} and the weight of the additional payload.

\begin{figure}
    \centering
    \resizebox{1.\linewidth}{!}{\input{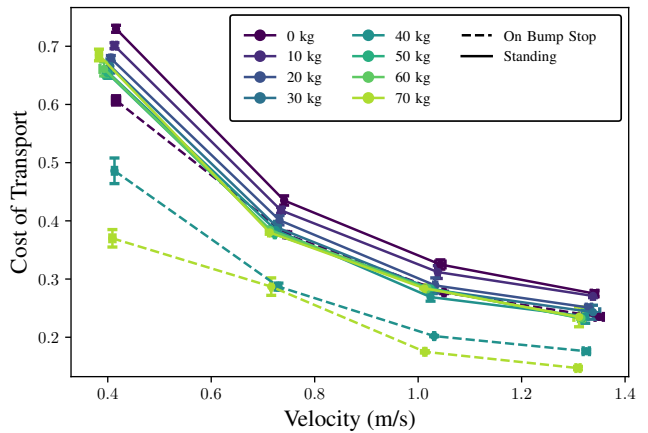}}  
    \caption{Plot showing \textit{LEVAs} \gls{cot} for different payload masses and comparing standing and sitting on bump stops.}
    \label{CoTtospeed}
\end{figure}

Our experiments involved four different constant velocity commands with varying weights from \SI{0}{\kilogram} to \SI{70}{\kilogram}, increasing in \SI{10}{\kilogram} increments. We recorded average velocities and power consumption over 10-meter stretches of flat ground and evaluated the \gls{cot} for these periods. The results can be seen in Fig. \ref{CoTtospeed}. This approach allowed us to calculate both the mean and standard deviation of \textit{LEVAs} \gls{cot} for each combination of weight and velocity. To minimize errors from border effects and acceleration, we excluded any data with velocities below a certain threshold of the commanded velocity.

A separate experiment was conducted in which \textit{LEVA} completely rests on its bump-stops, effectively not requiring the use of shoulder actuators. This was done to gather comparable data to wheeled-only robots. Here only payloads between 50 and \SI{70}{\kilogram} were evaluated. The smaller scale of this second test was chosen as the first tests showed no significant difference between payloads. The general test setup and evaluation process remained the same. 

The experimental data indicates:
\begin{itemize}
    \item The \gls{cot} tends to be inversely proportional to the velocity as expected.
    \item The bump-stops improve the \gls{cot} around 40\%.
    \item The variance in the measurements likely comes from slight steering inputs made by the human operator during the test.
\end{itemize}
Comparing the resulting \gls{cot} to robots in table \ref{table:CoTleg}, it can be noted that while \textit{LEVA} lacks behind wheeled systems, it outperforms systems with similar locomotion capabilities, such as quadrupeds.

\label{disc}
\section{Conclusion and Future Work}
\label{Conc}
In this paper, we introduced \textit{LEVA}, an innovative high-payload, high-mobility robot with hybrid locomotion capabilities designed for challenging terrains and high-efficiency payload transport. Initial tests have demonstrated the effectiveness of \textit{LEVA}'s unique configuration, which integrates both wheeled and legged mobility, showcasing an impressive payload capacity coupled with a remarkably low \gls{cot}\edited{on flat terrains}{}.

Notably, \textit{LEVAs} payload capacity closely aligns with specialized systems like Barry, which are specifically designed for efficient payload transport in similar operational contexts. When compared to purely wheeled industrial solutions, \textit{LEVA} currently does not achieve the same levels of efficiency, however, with its self-loading and unloading capabilities,
\textit{LEVA} has a distinct advantage compared to both legged as well as wheeled systems.

The successful box pickup experiments have demonstrated \textit{LEVAs} capability to autonomously handle payloads, emphasizing the reliability of this process. This ability, combined with an efficient wheeled and legged platform, could extend the reach of traditional logistics into unstructured terrains, where platforms similar to \textit{LEVA} could help address challenges in automation.

Future work will focus on refining \textit{LEVAs} energy efficiency by implementing springs, further robustifying locomotion control while transporting boxes on stairs, and enhancing its autonomy.

\bibliographystyle{IEEEtran}
\bibliography{refs}

\vfill

\end{document}